\documentclass[10pt,twocolumn,letterpaper]{article}

\usepackage{wacv}
\usepackage{times}
\usepackage{epsfig}
\usepackage{graphicx}
\usepackage{amsmath}
\usepackage{amssymb}

 \usepackage{subcaption}
\usepackage{mathrsfs}
\usepackage[ruled,vlined]{algorithm2e}

\def\wacvPaperID{} 

\wacvfinalcopy 

\ifwacvfinal
\def\assignedStartPage{1} 
\fi

\def\bd{\mathbf d}
\def\TD{\mathbf D}
\def\TK{\mathbf K}

\def\bp{\mathbf p}
\def\by{\mathbf y}

\def\bp{\mathbf p}

\def\bs{\mathbf s}
\def\bc{\mathbf c}

\def\bx{\mathbf x}
\def\TX{\mathbf X}
\def\TQ{\mathbf Q}
\newcommand{\grad}{\ensuremath{\nabla}}

\ifwacvfinal
\usepackage[breaklinks=true,bookmarks=false]{hyperref}
\else
\usepackage[pagebackref=true,breaklinks=true,colorlinks,bookmarks=false]{hyperref}
\fi

\ifwacvfinal
\else
\fi

\begin{document}

\title{Point-to-set distance functions for weakly supervised segmentation.}

\author{Bas Peters\\
Computational Geosciences Inc.\\
1623 West 2nd Ave, Vancouver, BC, Canada\\
{\tt\small {bas@compgeoinc.com}}}

\maketitle

\begin{abstract}
When pixel-level masks or partial annotations are not available for training neural networks for semantic segmentation, it is possible to use higher-level information in the form of bounding boxes, or image tags. In the imaging sciences, many applications do not have an object-background structure and bounding boxes are not available. Any available annotation typically comes from ground truth or domain experts. A direct way to train without masks is using prior knowledge on the size of objects/classes in the segmentation. We present a new algorithm to include such information via constraints on the network output, implemented via projection-based point-to-set distance functions. This type of distance functions always has the same functional form of the derivative, and avoids the need to adapt penalty functions to different constraints, as well as issues related to constraining properties typically associated with non-differentiable functions. Whereas object size information is known to enable object segmentation from bounding boxes from datasets with many general and medical images, we show that the applications extend to the imaging sciences where data represents indirect measurements, even in the case of single examples. We illustrate the capabilities in case of a) one or more classes do not have any annotation; b) there is no annotation at all; c) there are bounding boxes. We use data for hyperspectral time-lapse imaging, object segmentation in corrupted images, and sub-surface aquifer mapping from airborne-geophysical remote-sensing data. The examples verify that the developed methodology alleviates difficulties with annotating non-visual imagery for a range of experimental settings.
\end{abstract}

\section{Introduction}
Generating a large training set with fully segmented/annotated images for convolutional neural networks is costly and time consuming at best. Therefore, researchers developed methods for faster annotation and training algorithms to learn from the sparsely annotated images. Such methods include point annotations \cite{10.1007/978-3-319-46478-7_34,doi:10.1190/INT-2018-0225.1}, as well as annotated slices from a 3D data volume \cite{Unet3D}. More extreme cases to speed up the annotation time per image are bounding boxes \cite{Dai_2015_ICCV,rajchl2016deepcut,Khoreva_2017_CVPR,kervadec2020bounding} or knowledge if an object is present in the image (image-level supervision) e.g., \cite{10.1109/ICCV.2015.203,10.1007/978-3-319-46484-8_25}. Annotation time/cost of many categories of images is not a fundamental problem because most people can annotate street scenes or pictures of everyday objects and animals.

More serious challenges arise when just domain experts can annotate, for instance, hyperspectral and medical images. Yet a more difficult situation occurs when even to domain experts cannot annotate, and labels come from sparsely available ground truth. Examples include data that is not imaged yet, such as multi-modality geophysical data or corrupted images.

Many approaches for learning from image-level tags and bounding boxes use more involved workflows, with either multiple networks used in sequence, networks with multiple branches, or alternating between network training and updating estimates of pixel-level masks/region proposals. The typical goal is to reduce the annotation cost for datasets with large numbers of images. The scope of this work is different. We address the limitations of working with limited annotation in case labels come from ground truth or domain experts. Such data includes hyperspectral data, geophysical data, medical images, highly corrupted images, and indirect measurements that typically require solving an inverse problem to create an image. Many of the aforementioned datasets contain just one or a few examples.

In this work, we introduce a new method to obtain pixel-level segmentation from image-level information. Our algorithm can train in various modes on datasets as small as a single example: without any annotation, with partial annotation where the annotation is missing for one or more of the classes, and with bounding boxes. We inject image-level supervision directly into the learning problem via constraints on the network output via a new implementation based on point-to-set distance functions. Contrary to some statements in the literature about training networks with constraints on the output, we show that our approach does not lead to a very challenging or computationally expensive optimization problem. The distance functions, combined with a closer look at the optimization problem and the corresponding Lagrangian, reveal that the coupling between network parameters, network output, and constraints is not as complicated as it seems from a high-level problem formulation.

\subsection{Related work}
Constraints to the output of a neural network dates back to at least \cite{10.5555/2969644.2969708}. That work introduces (in)equality constraints encoded via differentiable functions as penalties or Lagrangian multipliers. The scope and applications are different from image segmentation. Various applications aimed at natural language, some not in the context of neural networks, use a probabilistic learning framework to introduce posterior constraints \cite{JMLR:v11:ganchev10a}, see also, e.g., \cite{10.5555/1795114.1795120, JMLR:v11:mann10a} for related work. \cite{JMLR:v11:ganchev10a} applies constraints that hold in expectation applied to multiple examples. \cite{NIPS2019_9385} use an alternating optimization method for differentiable non-linear constraints.

Various works on vision applications are closer to our work and include penalties/constraints that promote the presence or minimum/maximum size of objects \cite{10.1109/ICCV.2015.203,Pathak_2015_ICCV,marquez2017imposing,KERVADEC201988}. \cite{Pathak_2015_ICCV} implement linear inequality constraints on the network output with slack via an alternating optimization strategy by introducing auxiliary variables to remove direct coupling between the network output and the constraints. \cite{marquez2017imposing} present a Lagrangian method for training with non-linear but differentiable constraints, primarily for human-pose estimation. \cite{KERVADEC201988} and \cite{kervadec2020bounding} add a penalty or log-barrier on the violation of linear inequalities respectively. The aforementioned works all require differentiability of the penalty/constraint functions, and if provided, the gradient is specific to the particular penalty or constraint. These properties are specifics and limitations that we avoid using projection-based functions that measure the distance to constraint sets.

\subsection{Contributions}
The primary focus is on constraints applied to training a network on a single-example dataset in a visual context where there is no clear The primary focus and difference compared to related work from the previous section is the optimization: writing down the Lagrangian for training a network with constraints on the output shows that introducing auxiliary variables is unnecessary to obtain a simple optimization scheme without computationally expensive inner loops. A second difference is that we use the projection operator onto the constraint set directly in a distance function and its derivative. This enables the use of more than just linear inequalities or differentiable regularization functions. The projection-based approach means that different constraints change the projection operator, but the implementation, objective, and optimization all keep the same structure. Therefore, our approach easily extends to multiple types of constraints, including non-trivial intersections and sums of sets. We focus on applications that are single-example datasets in a visual context where there is no clear distinction between object and background. 
The contributions of this work are summarized as follows:

\begin{itemize}
\item Introduce a new distance-function based formulation and optimization scheme to include high-level image information as constraints while training neural networks for semantic segmentation.
\item Point-to-set distance function and its derivative use the projection operator, there is no need to define custom regularization functions and their derivatives. This approach also avoids challenges when the functions that describe constraints are not differentiable but it is still known how to project onto the corresponding set.
\item We extend the range of applications of constrained network training to imaging sciences problems without clear object-background setting, problems where annotating requires domain-experts or ground-truth observations, and single image/data problems.
\item The method applies to settings with \emph{a}) no annotation; \emph{b}) bounding boxes; \emph{c}) partial annotation where one or more classes have no annotation.
\end{itemize}

The organization of the remainder of this paper is as follows: we propose our approach conceptually. Next, we introduce a novel optimization implementation for training neural networks with constraints on the output, which does not rely on nested or alternating optimization schemes. We illustrate our contributions on three different problems from the imaging sciences. Finally, we discuss a few extensions of this work not covered in this paper.

\section{High-level information as constraints on the network output}

When fully annotated masks are not available, we can resort to high-level image information. In this work, we do not use image class-tags, but focus on quantitative prior-knowledge, particularly in terms of area/volume a certain class is expected to occupy in the segmentation. \cite{Pathak_2015_ICCV,KERVADEC201988} showed that such information is sufficient to obtain pixel-level segmentations for object detection tasks in both general and medical image datasets, while the prior knowledge does not need to be very precise.

In this work, we show that size information is also useful for datasets with just a single example, as well as for applications where there is no clear foreground/background or object. For instance, in hyperspectral imaging, we may know roughly the surface area of farm fields that changed their land use. Alternatively, unreliable manual annotations may provide upper and lower bounds on the surface area for a class. Other examples that do not have a background-object structure include geological/hydrological mapping from multi-modality geophysical and remote sensing data. In the following, we assume no knowledge of the spatial distribution of the class in the segmentation. While the approach outlined in this section can benefit from bounding boxes, we do not rely on them.

We now formalize the previous informal description, using for simplicity, two classes. One way to express the prior knowledge from the previous paragraph is via constraints on the area of a class in the segmentation
\begin{align}\label{prob_words1}
a_1 \leq &\text{Area(class 1)} \leq a_2 \\
(1-a_2) \leq &\text{Area(class 2)} \leq (1-a_1),
\end{align}
where $a_1$ and $a_2$ are scalar bounds. In 3D, the same holds in terms of volume. The goal is to obtain a segmentation that honours \eqref{prob_words1} by adding this information as constraints on the output of a neural network. Next, we present a new implementation and algorithm, which loosely speaking, can be considered as a generalization of work by \cite{Pathak_2015_ICCV,KERVADEC201988}.

\subsection{Constraints on the size/surface area of a class}
Denote 2D/3D/4D data in vectorized form as $\bd \in \mathbb{R}^{n_1 n_2 n_3 n_\text{chan}}$ (for 3D data input), with multiple space/time/frequency coordinates, one channel coordinate, and $N=n_1 n_2 n_3$. The non-linear function $g(\TK,\bd) : \mathbb{R}^{N n_\text{chan}} \rightarrow  \mathbb{R}^{N n_\text{class}}$ denotes a neural network for semantic segmentation that transforms the input data into $n_\text{class}$ probability maps with $N$ elements each. The network depends on parameters $\TK$, such as convolutional kernels. The vector $\bc$ contains the labels, which may be partial or not available at all for one or more classes. 

The main algorithmic contributions of this work are a more direct way to include surface area/size information and a new implementation to train neural networks with constraints on the output. \\
\textbf{Cardinality constraints:} The cardinality of a vector, $\operatorname{card}(\bx)$, counts the number of non-zero elements in $\bx$. Therefore, we have a direct translation from \eqref{prob_words1} as the sets
\begin{align}\label{const_sets1}
&\mathcal{C}_1 \equiv \{ g(\TK,\bd) \: | \: \operatorname{card}(g(\TK,\bd)_1) \leq a_2 \} \\
&\mathcal{C}_2 \equiv \{ g(\TK,\bd) \: | \: \operatorname{card}(g(\TK,\bd)_2) \leq (1-a_1) \}, 
\end{align}
where $g(\TK,\bd)_1$ indicates the first channel of the network output $g(\TK,\bd)$. We assumed that the last layer of the network normalizes the output to $0-1$ and sums to one, for example, via the softmax function. This assumption allows us to use upper bounds on the cardinality only. Lower bounds are implicit via the normalization of the output, i.e., if $\operatorname{card}(g(\TK,\bd)_1) \leq a_2$, it implies that $(1-a_2) \leq\operatorname{card}(g(\TK,\bd)_2)$. The projection onto the set of vectors with limited cardinality is known in closed form by setting all but the largest values (in absolute sense) to zero. A typical implementation is: sort the vector, determine the threshold value, set all other elements to zero, revert to original order. 

\subsection{Point-to-set distance functions for networks with output-constraints}
If the only available information is prior knowledge on some property of the output of a network, we seek to solve the feasibility problem
\begin{equation} \label{feas_prob}
\operatorname{find} g(\TK,\bd) \in \mathcal{D} \Leftrightarrow \min_{\{\TK\}} \iota_\mathcal{D}(g(\TK,\bd)),
\end{equation}
where $\iota_\mathcal{D}$ is the indicator function for the set $\mathcal{D}$. If partial annotation for at least one of the classes is available, we add a loss function for the labels, $l$, (e.g., cross-entropy) and optimize over network parameters $\TK$ subject to constraints:
\begin{equation} \label{prob_const}
\min_{\TK} \:\:  l(\TQ g(\TK,\bd),\bc) \:\: \text{s.t.} \:\: g(\TK,\bd) \in \mathcal{D},
\end{equation}
where the matrix $\TQ$ selects from the output the pixels where labels, $\bc$, are available. To be more general and allow multiple constraint sets simultaneously, we use the intersection of $p$ sets
\begin{equation}
\mathcal{D} \equiv \bigcap_{i=1}^p \mathcal{C}_i \notin \emptyset,
\end{equation}
which should be non-empty. The training of neural networks typically relies on variants of stochastic gradient descent to minimize the loss. The most obvious extension to problem \eqref{prob_const} may seem the project (stochastic) gradient descent iterations $\TK^{i+1/2} = \TK^i - \gamma \grad_{\TK} l(\TQ g(\TK_i,\bd),\bc)$ with stepsize $\gamma$, followed by the Euclidean projection such that the network output $g(\TK,\bd)$ is an element of the set $\mathcal{D}$, i.e., $\TK^{i+1/2} \in \arg\min_{\TK^{i+1}} \frac{1}{2} \| \TK^{i+1} - \TK^{i+1/2} \|_2^2 \:\: \text{s.t.} \:\: g(\TK^{i+1},\bd) \in \mathcal{D}$. Because the constraints act on the network output and not on the parameters over which we optimize directly, this is not a straightforward projection problem. To proceed, we could introduce auxiliary variables and equality constraints to construct a projection problem that requires $g(\TK^{i+1},\bd)$ to be an element of $\mathcal{D}$ indirectly, see, e.g., \cite{Pathak_2015_ICCV}.

In this work we present a new implementation of the constrained problems \eqref{prob_const} and \eqref{feas_prob} that is simpler in the sense that we stay close to the original problem formulation \eqref{prob_const}, and there are no auxiliary variables or computationally expensive alternating optimization schemes. The point-to-set distance function is at the core of our approach. We use a version that measures the squared distance from a point in $\mathbb{R}^N$ (vector $\by$) to the set $\mathcal{D}$,
\begin{equation}\label{distsq}
d^2_\mathcal{D}(\mathbf{y}) = \min_{x \in  \mathcal{D}} \frac{1}{2}\| \bx - \by \|_2^2 = \frac{1}{2} \|  \mathcal{P}_\mathcal{D}(\by) - \by \|_2^2
\end{equation}
using just the projection operator of a vector onto the intersection of constraints, $\mathcal{P}_\mathcal{D}$. We could also opt for an exact version by removing the square \cite[Thm. 1.2.3]{hiriart1996convex} but this is not of paramount importance for our applications. The squared distance function is differentiable \cite[Ex. 3.3]{hiriart2004fundamentals} as
\begin{equation}\label{distsqgrad}
\nabla_{\by} d^2_\mathcal{D}(\mathbf{y}) = \by -  \mathcal{P}_\mathcal{D}(\by)
\end{equation}
This is expression holds, even if the constraint set corresponds to a non-differentiable regularizer. A closed-form derivative of the distance function using the projection operator itself is a powerful result that is also at the core of algorithms for (split-) feasibility problems (e.g., the CQ algorithm \cite{Byrne_2002}). Note that the functional form of the constraint, or the associated scalar/vector upper/lower bounds do not appear in the expression of the distance function and its derivative; they are implicit in the projection operation. This property makes the distance function a general tool that applies to any constraint set, including intersections.

We replace the constraint on the network output (the final network state $g(\TK,\bd)=\by_n$) with the squared distance-function. We thus minimize the distance to the constraint set, plus the loss related to the labels (if any), subject to one equality constraint per neural network layer,
\begin{align}\label{prob}
\min_{\{\TK\}} \:\: & l(\TQ \by_n,\bc) + \frac{\alpha}{2}  \|  \mathcal{P}_\mathcal{D}(\by_n) -  \by_n \|_2^2 \:\: \text{s.t.} \\
&\by_n = \by_{n-1} - f(\TK_n \by_{n-1}) \nonumber \\
&\vdots \nonumber \\
&\by_j = \by_{j-1} - f(\TK_j \by_{j-1}) \nonumber  \\
&\vdots \nonumber \\
&\by_1 = \bd, \nonumber
\end{align}
where the network state at the first layer is equal to the data $\bd$ and the scaling parameter $\alpha>0$. Problem \eqref{prob} is not a standard penalty approximation to a constrained problem because the distance function has a number of advantages over standard penalty functions, which we discuss and exploit in the following section. In the above problem statement, we used a standard ResNet \cite{he2016deep} with a nonlinear activation function $f(\cdot)$ to keep notation compact. We emphasize that this work does not depend on the ResNet, and most neural networks could replace the ResNet in the above problem statement. Below, we show that minimizing a distance function applied to the output of a network does not fundamentally change the optimization process for training a neural network based on labels alone. The proposed approach combines seamless with most existing neural networks and their training algorithms. To show this, consider the Lagrangian corresponding to \eqref{prob},

\begin{align}\label{Lag}
&L(\{\by\},\{\bp\},\{\TK\}) = l(\TQ \by_n,\bc) + \frac{\alpha}{2}  \|  \mathcal{P}_\mathcal{D}(\by_n) -  \by_n \|_2^2 \\
- &\sum_{j=2}^n  \bp_j^\top (\by_j - \by_{j-1} + f(\TK_j \by_{j-1})) -  \bp_1^\top (\by_1 - \bd) \nonumber ,
\end{align}
where $\bp_j$ are the vectors of Lagrangian multipliers for every layer $j$. For optimization, we need the following partial derivatives of the Lagrangian:

\begin{align}\label{LagGrads}
&\grad_{\by_n} L = \grad_{\by_n} l(\TQ \by_n,\bc) + \alpha (\by_n -  \mathcal{P}_\mathcal{D}(\by_n)) - \bp_n\\
&\nonumber\\
&\text{for} \: j=n-1,\cdots,2 : \nonumber\\
&\grad_{\by_{j-1}} L  = -\bp_{j-1} + \bp_j -\TK_{j-1}^\top  \operatorname{diag}(f'(\TK_{j-1} \by_{j-1}))  \bp_{j} \\
&\grad_{\bp_j} L = -\by_j + \by_{j-1} - f(\TK_j \by_{j-1})\\
&\grad_{\TK_j} L = \bigg[ \frac{\partial \big(\ \TK_j \by_{j-1} \big)}{\partial \TK_j}\bigg]^\top \operatorname{diag}(f'(\TK_j \by_{j-1})) \bp_j.
\end{align}
The function $f'(\cdot)$ is the derivative of the activation and the operator $\operatorname{diag}(\cdot)$ creates a diagonal of a matrix of the input. The gradient updates for the network parameters $\TK_j$ at every layer follow via the backpropagation algorithm. First, forward propagating through the network will satisfy all equality constraints and thus $\grad_{\bp_j} L = 0$. Second, propagating backwards provides the Lagrangian multipliers and $\grad_{\by_{j}} L =0$. The last step computes the gradient with respect to the network parameters using the already computed quantities. Algorithm \ref{alg:backprop} summarizes these steps.

Compared to training using labels only, is that the constraints insert information into the final Lagrangian multiplier that then backpropagates through all layers.

\subsection{Stopping criteria and choice of $\alpha$}
In this work, we assume either \emph{a}) no annotated data; \emph{b}) one or more classes come without any annotation; \emph{c}) bounding boxes.

\textbf{Case (a)} Because there are no labels, the loss reduces to the point-to-set distance function for finding a feasible point \eqref{feas_prob}. We stop training when the output of the network, $\by_n = g(\TK,\bd)$, is an element of the constraint set $\mathcal{D}$, i.e., $\by_n  \in \mathbf{D}$. The implementation via an inexact penalty function (problem \eqref{prob}) means we need to increase the penalty parameter $\alpha$ until we achieve $\| \by_n - \mathcal{P}_\mathcal{D}(\by_n) \| \leq \varepsilon$.

\textbf{Case (b \& c)}
(\emph{c}) is a special case of (\emph{b}). As there are no labels for one or more classes, we cannot use standard early stopping (saving best model parameters) at the lowest validation loss. The constraints are all information we have on the classes without annotation, thus we need to look at the lowest validation loss for the classes with annotation, while satisfying the constraints within an $\epsilon$-tolerance as above. 

Algorithm \ref{alg:backprop} summarizes the workflow to train a network subject to constraints on the output \eqref{prob_const}. For the feasibility problem \eqref{feas_prob} we set $\alpha=1$. For the label-loss plus point-to-set distance \eqref{prob_const}, we increase the penalty parameter $\alpha$ if the distance to the constraint set does not decrease in a window of a few iterations.
\begin{algorithm}[]
\SetAlgoLined
$\by_{1} = \bd$, $\alpha>0$ (penalty parameter), $\eta>1$ (penalty growth factor), $m$ (history length), $\gamma$ (learning rate)\; 
\For{$i=1:n_\text{max iter}$}{
 \For{$j=2,\cdots,n$}{
  $\by_j = \by_{j-1} - f(\TK_j \by_{j-1}) $ // Forward\;
 }
 $\bp_{n} = \grad_{\by_n} l(\TQ \by_n  , \bd) + \alpha (\by_n -  \mathcal{P}_\mathcal{D}(\by_n))$ //Final Lagrangian multiplier\;
 //Propagate backward and update network parameters for each layer\;
 \For{$j=n,n-1,\cdots,2$}{
 $\grad_{\TK_j} L = \bigg[ \frac{\partial \big(\ \TK_j \by_{j-1} \big)}{\partial \TK_j}\bigg]^\top \operatorname{diag}(f'(\TK_j \by_{j-1}) \bp_j$ \;
 $\TK_j \leftarrow \TK_j - \gamma \grad \TK_j$ \;
  $ \bp_{j-1} = \bp_j -\TK_{j-1}^\top  \operatorname{diag}(f'(\TK_{j-1} \by_{j-1})) \bp_{j}$\;
}
$d_i = \|\ \by_n -  \mathcal{P}_\mathcal{D}(\by_n) \|_2^2$ \;
\If{$\text{max}\{d_{i-1}, d_{i-2},\cdots,d_{i-m}\} < d_i $}{
$\alpha \leftarrow \alpha \eta$ //Update penalty parameter
}
}
 \caption{Backpropagation to train a network including constraints on the network output via distance-to-set functions.}
\label{alg:backprop}
\end{algorithm}

\section{Examples}
The emphasis of the examples is on data from the imaging sciences, where we often have a single example with annotation that comes from domain experts or ground truth. Therefore, we are explicitly not interested in common datasets with many examples that are easy to annotate, such as MS-COCO or PASCAL VOC datasets. While the previous section is guided by the notationally compact ResNet, the examples use a fully invertible (or reversible) hyperbolic network \cite{Chang2017Reversible,lensink2019fully}. Invertibility removes the requirement to store all network states for gradient computations. As a result, a reversible network can train on large scale data that would otherwise not fit on a standard GPU. Examples of such data include time-lapse hyperspectral, and data with a large number of channels like the remote sensing/geoscience example in the following section. The reversible hyperbolic network follows the recursion $\by_{j} = 2\by_{j-1} -  \by_{j-2} -  h^2 \TK_j^\top f( \TK_j \by_{j-1})$, which shows that the network structure does not alter Algorithm \ref{alg:backprop} fundamentally because $\grad_{\by_n} L$ still leads to a closed-form solution for $\bp_n$. All experiments use the timestep $h=0.2$. The network for the following hyperspectral example is $24$ layers deep with $12\times12$ convolutional kernels per layer; ten layers for the single image segmentation example with $9\times9$ convolutional kernels; and the multi-modality sub-surface characterization example has a network with $34$ layers. For this last example, there are $53$ input channels but $\TK$ has a `flat' block structure of $12 \times 53$ convolutional kernels to limit the number of free parameters and induce a block-low-rank structure in $\TK_j^\top f( \TK_j )$ \cite{peters2019symmetric}.

\subsection{No labels or bounding boxes: time-lapse hyperspectral land-use change detection}
\begin{figure*}[!htb]
 	\centering
 	\begin{subfigure}[b]{0.27\textwidth}
 		\includegraphics[width=\textwidth]{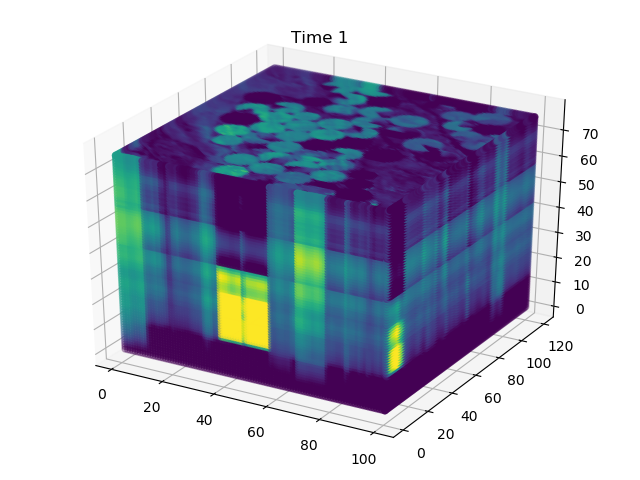}
 		\caption{}
 		\label{fig:Figure1a}
 	\end{subfigure}
 	\begin{subfigure}[b]{0.27\textwidth}
 		\includegraphics[width=\textwidth]{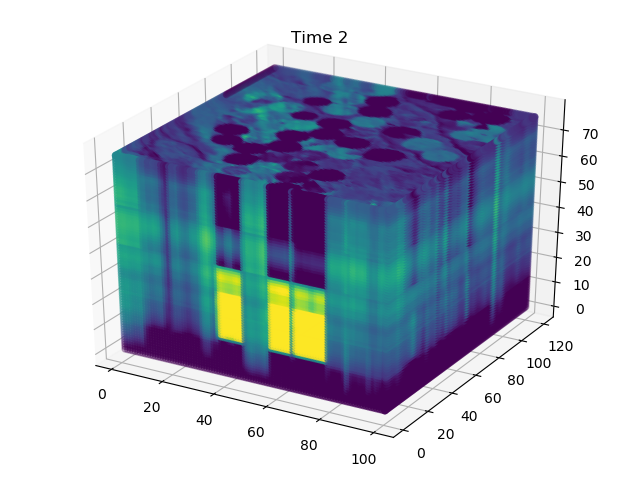}
 		\caption{}
 		\label{fig:Figure1b}
 	\end{subfigure}
 	 	\begin{subfigure}[b]{0.22\textwidth}
 		\includegraphics[width=\textwidth]{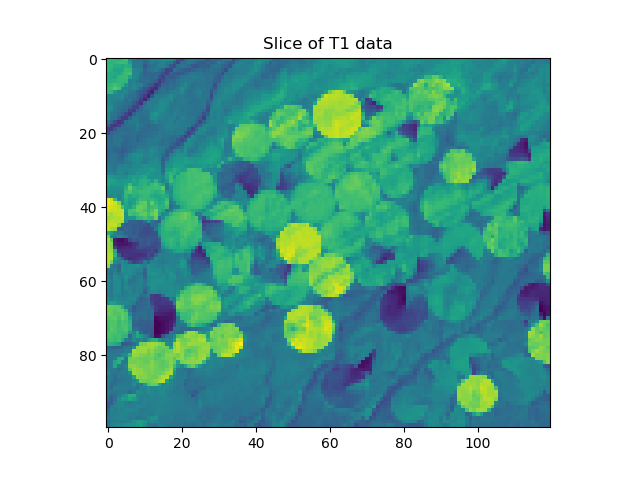}
 		\caption{}
 		\label{fig:Figure1c}
 	\end{subfigure}
 	 	 	\begin{subfigure}[b]{0.22\textwidth}
 		\includegraphics[width=\textwidth]{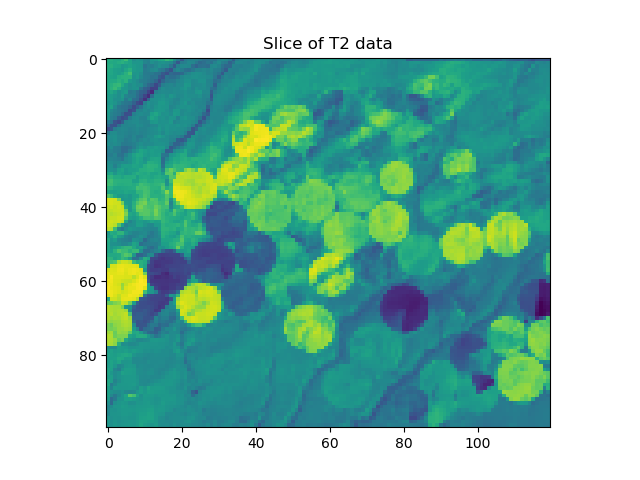}
 		\caption{}
 		\label{fig:Figure1d}
 	\end{subfigure}
 	\caption{(a) and (b): Two hyperspectral datasets recorded at different times for time-lapse land-use change detection. (c) and (d): slices from (a) and (b) for a single frequency.}
\label{fig:HypData}
 \end{figure*}

The goal of time-lapse hyperspectral land-use change detection is to create a 2D change map of a piece of the earth from two 3D hyperspectral datasets, collected at different times \cite{doi:10.1080/01431161.2018.1466079}, see Figure \ref{fig:HypData}. Domain-experts or ground truth can provide annotation. However, this is expensive, time-consuming, and prone to errors. 

Working with just one example without any annotation, we can obtain reasonably accurate predictions using knowledge on the surface area occupied by the two classes (no-change / change). Such prior knowledge may be available from sources like historical estimates or unreliably labeled data that still provides bounds on the surface area.

The true surface area for land-use change is $\approx 34\%$. We use loose bounds $25\% \leq \text{Area(class 1)} \leq 45\%$. This translates to the cardinality constraints \eqref{const_sets1} with $a_1 = 0.45$ and $a_2=0.75$. We employ gradient descent as in Algorithm \ref{alg:backprop} to find a solution to the feasibility problem \eqref{feas_prob}. To induce stochasticity when working with a single example, we apply random flips and permutations to the data at every iteration.

\begin{figure*}[!htb]
 	\centering
 		\includegraphics[width=0.8\textwidth]{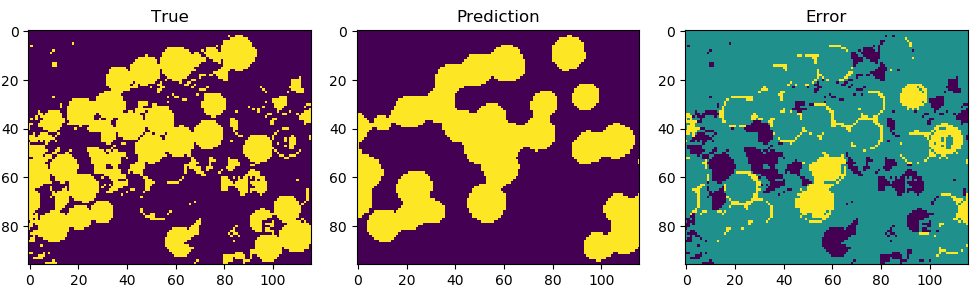}
 	\caption{The provided annotations (not used for training), our prediction that does not use any labels, and the difference. Most of the errors are boundary effects, as well as a few false positive/negative identified fields.}
\label{fig:HypResults}
 \end{figure*}

Figure \ref{fig:HypResults} shows the provided annotations (not used in this example), our prediction, and the difference. The most `obvious' segmentation would be \emph{all} farm fields as one class. The results, however, show a prediction that mostly selects the subset of the fields that changed land use. Except for a few errors, our method was able to capture the subtle differences over time. While these results are not as good as \cite{FRHyperspectral}, we did not require any annotation.

\subsection{Point annotations for one of the classes only: Multi-modality subsurface characterization}

\begin{figure*}[!htb]
 	\centering
 		\includegraphics[width=0.9\textwidth]{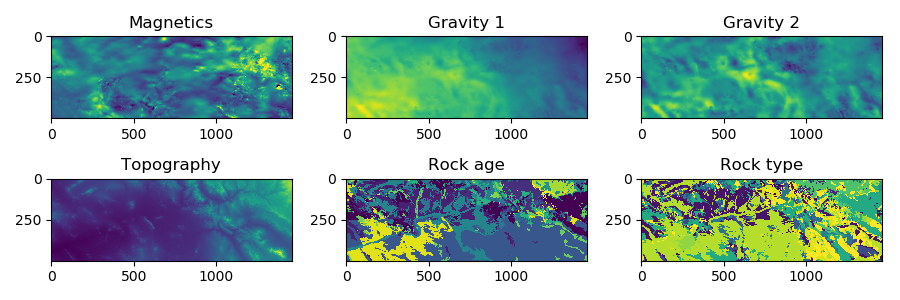}
 	\caption{Data for the multi-modality remote sensing example. The rock age and rock type maps are converted to $53$ separate maps that indicate if the rock age/type is present or not.}
\label{fig:AqData}
 \end{figure*}

\begin{figure*}[!htb]
 	\centering
 		\includegraphics[width=\textwidth]{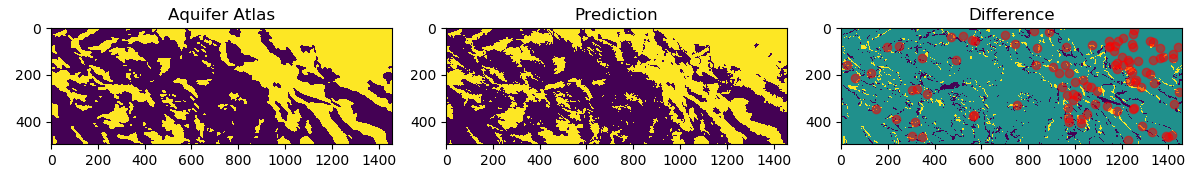}
 	\caption{Left: aquifer map, based on data and manual interpretation. Middle: prediction. Right: difference between prediction and full map. Red dots represent the annotation. Note that there is annotation for one of the two classes only.}
\label{fig:AqResults}
 \end{figure*}

Figure \ref{fig:AqData} shows various types of remote/airborne sensing data and geological maps. The goal is to obtain a map of the aquifers (yes/no aquifer present in the subsurface) in an area that is roughly $40 \%$ of Arizona, see Figure \ref{fig:AqResults}. The labels from \cite{wateratlas} are a combination of ground-truth observations, remote sensing, geophysical data, and expert interpretation. The goal is to see if we can reproduce domain-experts' work, based on $100$ point annotations for only one of the two classes, supplemented with an estimate of the total area of aquifers being between $50\%$ and $65\%$. For some applications, it is easier to annotate either true or false for a specific question. The prediction in Figure \ref{fig:AqResults} shows we can reproduce domain-experts interpretation using point annotations from one of the two classes. The differences are minor along the boundaries, as well as a few small patches.

 \subsection{Segmenting a corrupted image using a bounding box}
\begin{figure*}[!htb]
 	\centering
  	 	\begin{subfigure}[b]{0.33\textwidth}
 		\includegraphics[width=\textwidth]{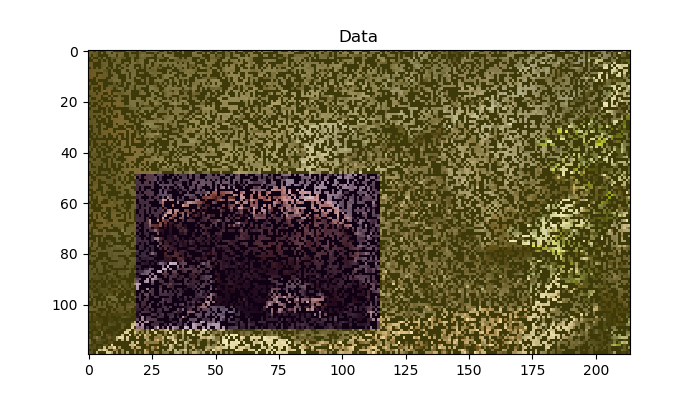}
 		\caption{}
 		\label{fig:Figure3a}
 	\end{subfigure}
 	 	 	\begin{subfigure}[b]{0.33\textwidth}
 		\includegraphics[width=\textwidth]{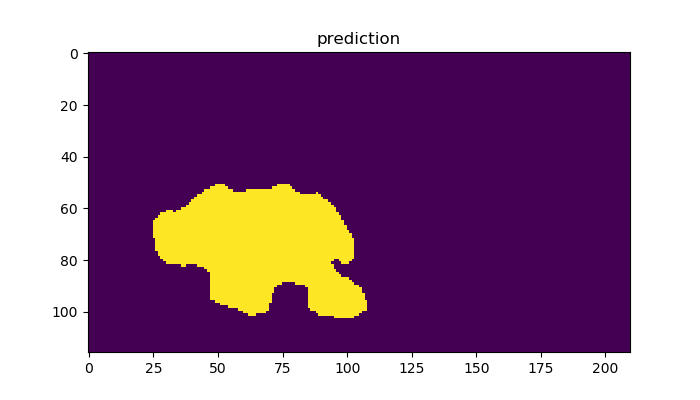}
 		\caption{}
 		\label{fig:Figure3b}
 	\end{subfigure}
 	 \begin{subfigure}[b]{0.33\textwidth}
 		\includegraphics[width=\textwidth]{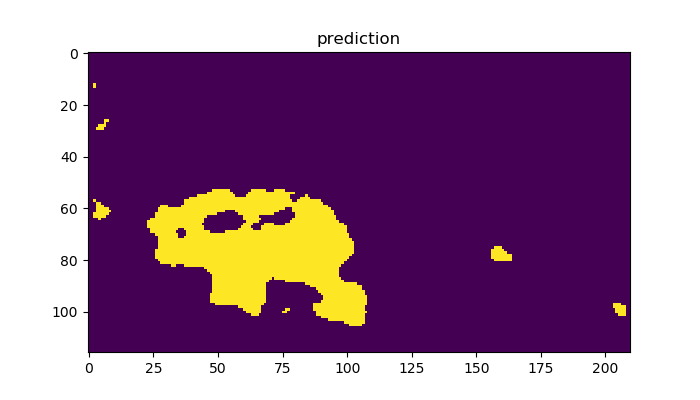}
 		\caption{}
 		\label{fig:Figure3c}
 	\end{subfigure}
 	\caption{(a) Corrupted image with $50\%$ missing pixels with highlighted bounding box; (b) the segmentation obtained using cardinality constraints on object size derived from the bounding box. Background and anomaly have an intersection over union accuracy of $98.0\%$ and $87.0\%$ respectively. (c) Comparison with a penalty related to a constraint on the sum of the network output, see section \ref{compsect} for details.}
 	\label{fig:SIBB2Pix_corrupted}
 	\end{figure*}

Our method can also provide a pixel-level segmentation of a single corrupted image using a bounding box. The data is an image from \cite{IEEEDavisDataset} with $50\%$ missing pixels. Manually generating pixel-level annotations would be challenging in this case. This task is different from most standard tasks that train on entire datasets and aim to generalize to unseen images. Figure \ref{fig:SIBB2Pix_corrupted} displays the target image and bounding box. From the box, we derive the constraint $\operatorname{card}(g(\TK,\bd))_1 \leq 23\%$ for the object class, and we set the constraint for the background class to $\operatorname{card}(g(\TK,\bd))_2 \leq 90\%$. The bounding box also implies we use the area outside the box as labels for the background class. There are no labels for the object class. The prediction in Figure \ref{fig:SIBB2Pix_corrupted} shows that constraints on the network output help obtain an accurate pixel-level segmentation for a single image from a bounding box.

The network was trained using stochastic gradient descent for 400 iterations. At each iteration, we induce randomness by randomly sampling $10\%$ of the background labels and randomly flipping and permuting the data.

\subsection{Comparison to other approaches}\label{compsect}
Because the scope of this work is either size information,
bounding boxes, or missing (partial) annotation one of the
classes with size information, we do not compare to methods that are specific to learning from bounding
boxes. Furthermore, comparisons are meaningful only if the method can also deal with non-visual multi-modality data
that we use in the following examples. 

We compare our approach with linear inequality constraint of the form $a_1 \leq \sum_{i=i}^{N} g(\TK,\bd)_{1,i} \leq a_2$. Bounds on the sum of do not directly tell us how many pixels will be classified as a certain class because $\sum_{i=i}^{N} g(\TK,\bd)_{1,i}$ can be concentrated in a minimal number of pixels, or be spread out evenly among a large number of pixels. Specifically, we show results using the penalty used by \cite{KERVADEC201988}, which reads $+\lambda (\mathbf{1}^\top \bs - a_1)^2$ if $\mathbf{1}^\top \bs < a_1$ and $+\lambda (\mathbf{1}^\top \bs - a_2)^2$ if $ \mathbf{1}^\top \bs > a_2$, where we used $\bs = g(\TK,\bd)_{1}$ and $\mathbf{1}$ is the vector of ones to keep notation compact. The gradient for this penalty is available in closed form as $\mathbf{1}(\mathbf{1}^\top \bs - a_1)$ and similarly for the other case. This shows the gradient is a constant shift of all parameters, which is very different from the gradient of the squared distance function, $\nabla_{\by} d^2_\mathcal{D}(\mathbf{y}) = \by -  \mathcal{P}_\mathcal{D}(\by)$, where $\mathcal{P}_\mathcal{D}$ projects onto the set of vectors with limited cardinality that shifts most elements to zero, but leaves other elements untouched. This brief comparison also shows that the penalty function described above is specific for this type of constraints, as is the corresponding gradient. In our distance-function based approach, we only need to change $\mathcal{P}_\mathcal{D}$ to use different constraints.

Figure \ref{fig:SIBB2Pix_corrupted} shows a comparison of the proposed method with the penalty described above. We show the result for the best $\lambda$, selected by manual tuning. This result is not comprehensive but shows that the task is not trivial for any approach that includes information on the area a given class should occupy in the segmentation. The primary goal of our work is to introduce a novel algorithm to implement constraints on the network output. Extensive numerical comparisons are secondary and beyond the scope of this paper.

\section{Computational cost}
The added computational cost of the gradient of the squared distance function is one projection per network output-channel per example. Practical timing depends on the availability of a GPU implementation for the projection, in order to avoid slower CPU computations and loading from and to the GPU. We implement the projection for the cardinality constraint using \emph{1}) sort; \emph{2}) thresholding most elements of the vector to zero; \emph{3}) permute back to the original ordering. The relative added computational time is low when the forward propagation is expensive because of a deep network or large input data size.
 	
\section{Extensions}
This section contains some extensions that are readily available using the presented material, and do not require additional computational tools or implementations.

\textbf{Other constraints for size information}.
Besides the cardinality constraints, there are at least two ways to represent size/surface area information. First, consider the convex relaxation of the cardinality function: the $\ell_1$ norm: $\mathcal{C} \equiv \{ g(\TK,\bd) \: | \: \| g(\TK,\bd)_1 \|_1 \leq a_2 \}$. A disadvantage of this constraint is that any network output within the $\ell_1$-ball satisfies the constraints without necessarily telling us anything about how many pixels/voxels are classified as the desired class. Yet another implementation of size constraints on the network output is via histogram constraints. 

\textbf{Structural constraints}.
If there is information about the length of the boundary of a given class in the segmentation, we can represent this using the cardinality of the spatial derivative of the network output, i.e., $\mathcal{C} \equiv \{ \bx \: | \: \operatorname{card}(\TD \bx) \leq k\}$, where $k$ is an interger, and $\TD$ is a discrete representation of the spatial derivatives of the network output. Other types of structural information derives from the observation that segmentations are often `simplified' versions of input images. In this case, the data provides bounds on the complexity of the spatial structure of the segmentation. For instance, the rank of the segmentation should be at most the rank of the input image, leading to the constraint $\mathcal{C} \equiv \{ \bx \: | \: \operatorname{rank}(\TX) \leq k\}$, where $\TX$ is the network output in matrix form, and $k$ is the desired maximum rank.

\textbf{Intersection of multiple constraint sets}.
Within our frameqork, there are at least two ways to include multiple constraints simultaneously. \emph{a}) using tools similar to \cite{Censor_2005}, we can add multiple distance penalties $+\alpha_1 / 2 \| \mathcal{P}_{\mathcal{C}_1}(\bx) - \bx \|_2^2 + \alpha_2 / 2 \| \mathcal{P}_{\mathcal{C}_2}(\bx) - \bx \|_2^2 + \cdots$. This implementation is straightforward, but introduces multiple penalty parameters that need tuning to make sure the solution is an element of the intersection. \emph{b}) a single distance function that includes the projection operator onto the intersection, $\mathcal{P}_{\mathcal{C}_1 \bigcap \mathcal{C}_2 \bigcap \cdots}$ avoids trade-off between multiple penalties: all constraints will be satisfied at the solution as long as the intersection is non-empty. Specialized software is available to compute projection onto intersections of sets \cite{peters2019algorithms}.

\textbf{Unsupervised data exploration}.
If there is no annotation and no prior knowledge to define any constraints, we can still solve the feasibility problem \eqref{feas_prob} for unsupervised deep clustering. Segmenting the input data multiple times, each time using different bounds on the size of each class, may reveal interesting patterns.

\textbf{Maintaining feasibility while training.}
Maintaining feasibility (`hard' constraints) is desired or required for some imaging problems \cite{herrmann2019learned}. To enforce feasibility while solving \eqref{prob}, we take multiple gradient steps based on the distance to the constraint set only, and discard the gradient w.r.t. the labels if the network output is not feasible.

\section{conclusions}
For data from the imaging sciences where annotation can only come from ground truth, or perhaps domain experts, the lack of annotation is the primary obstacle to obtain pixel-level segmentations. Examples include remote sensing data, geophysical data, corrupted images, or data is not `imaged' yet. Including prior knowledge on the minimum/maximum size/surface-area of an object proved to be a weak supervision technique that enables the segmentation of general and medical images using bounding boxes. We showed that such weak supervision also applies to data from the imaging sciences without clear object-background structure, even when only a single example is available. To directly implement class-size information, we use constraints on the cardinality of a vector, and introduced a novel training algorithm based on point-to-set distance functions. This approach requires the projection operator only, so that different constraints do not require translation into custom penalty functions and their derivatives. A quick look at the Lagrangian shows that the constraints implemented via a distance function fit seamless into training networks via backpropagation. Examples showed that constraints can replace missing annotation for a range of segmentation problems in the imaging sciences. Particularly, we showed segmentations from hyperspectral data without any annotation, segmentation of corrupted images from bounding box information, and segmentation from multi-modality remote sensing and geophysical data where we have sparse annotation for one of the classes only. 

{\small
\bibliographystyle{ieee_fullname}
\bibliography{biblio,VideoRefs,HyperSpectralRefs}

\begin{thebibliography}{10}\itemsep=-1pt

\bibitem{10.1007/978-3-319-46478-7_34}
Amy Bearman, Olga Russakovsky, Vittorio Ferrari, and Li Fei-Fei.
\newblock What's the point: Semantic segmentation with point supervision.
\newblock In Bastian Leibe, Jiri Matas, Nicu Sebe, and Max Welling, editors,
  {\em Computer Vision -- ECCV 2016}, pages 549--565, Cham, 2016. Springer
  International Publishing.

\bibitem{10.5555/1795114.1795120}
Kedar Bellare, Gregory Druck, and Andrew McCallum.
\newblock Alternating projections for learning with expectation constraints.
\newblock In {\em Proceedings of the Twenty-Fifth Conference on Uncertainty in
  Artificial Intelligence}, UAI ’09, page 43–50, Arlington, Virginia, USA,
  2009. AUAI Press.

\bibitem{Byrne_2002}
Charles Byrne.
\newblock Iterative oblique projection onto convex sets and the split
  feasibility problem.
\newblock {\em Inverse Problems}, 18(2):441--453, mar 2002.

\bibitem{Censor_2005}
Yair Censor, Tommy Elfving, Nirit Kopf, and Thomas Bortfeld.
\newblock The multiple-sets split feasibility problem and its applications for
  inverse problems.
\newblock {\em Inverse Problems}, 21(6):2071--2084, nov 2005.

\bibitem{Chang2017Reversible}
Bo Chang, Lili Meng, Eldad Haber, Lars Ruthotto, David Begert, and Elliot
  Holtham.
\newblock Reversible architectures for arbitrarily deep residual neural
  networks.
\newblock In {\em AAAI Conference on AI}, 2018.

\bibitem{Unet3D}
{\"O}zg{\"u}n {\c{C}}i{\c{c}}ek, Ahmed Abdulkadir, Soeren~S. Lienkamp, Thomas
  Brox, and Olaf Ronneberger.
\newblock 3d u-net: Learning dense volumetric segmentation from sparse
  annotation.
\newblock In Sebastien Ourselin, Leo Joskowicz, Mert~R. Sabuncu, Gozde Unal,
  and William Wells, editors, {\em Medical Image Computing and
  Computer-Assisted Intervention -- MICCAI 2016}, pages 424--432, Cham, 2016.
  Springer International Publishing.

\bibitem{Dai_2015_ICCV}
Jifeng Dai, Kaiming He, and Jian Sun.
\newblock Boxsup: Exploiting bounding boxes to supervise convolutional networks
  for semantic segmentation.
\newblock In {\em The IEEE International Conference on Computer Vision (ICCV)},
  December 2015.

\bibitem{JMLR:v11:ganchev10a}
Kuzman Ganchev, Jo{\~a}o Gra{\~A}, Jennifer Gillenwater, and Ben Taskar.
\newblock Posterior regularization for structured latent variable models.
\newblock {\em Journal of Machine Learning Research}, 11(67):2001--2049, 2010.

\bibitem{doi:10.1080/01431161.2018.1466079}
Mahdi Hasanlou and Seyd~Teymoor Seydi.
\newblock Hyperspectral change detection: an experimental comparative study.
\newblock {\em International Journal of Remote Sensing}, 39(20):7029--7083,
  2018.

\bibitem{he2016deep}
Kaiming He, Xiangyu Zhang, Shaoqing Ren, and Jian Sun.
\newblock Deep residual learning for image recognition.
\newblock In {\em Proceedings of the IEEE Conference on Computer Vision and
  Pattern Recognition}, pages 770--778, 2016.

\bibitem{herrmann2019learned}
Felix~J Herrmann, Ali Siahkoohi, and Gabrio Rizzuti.
\newblock Learned imaging with constraints and uncertainty quantification.
\newblock {\em arXiv preprint arXiv:1909.06473}, 2019.

\bibitem{hiriart1996convex}
Jean-Baptiste Hiriart-Urruty and Claude Lemarechal.
\newblock {\em Convex Analysis and Minimization Algorithms I: Fundamentals},
  volume 305.
\newblock Springer Science \& Business Media, 1996.

\bibitem{hiriart2004fundamentals}
Jean-Baptiste Hiriart-Urruty and Claude Lemar{\'e}chal.
\newblock {\em Fundamentals of Convex Analysis}.
\newblock Springer Science \& Business Media, 2004.

\bibitem{KERVADEC201988}
Hoel Kervadec, Jose Dolz, Meng Tang, Eric Granger, Yuri Boykov, and Ismail~[Ben
  Ayed].
\newblock Constrained-cnn losses for weakly supervised segmentation.
\newblock {\em Medical Image Analysis}, 54:88 -- 99, 2019.

\bibitem{kervadec2020bounding}
Hoel Kervadec, Jose Dolz, Shanshan Wang, Eric Granger, and Ismail ben Ayed.
\newblock Bounding boxes for weakly supervised segmentation: Global constraints
  get close to full supervision.
\newblock In {\em Medical Imaging with Deep Learning}, 2020.

\bibitem{Khoreva_2017_CVPR}
Anna Khoreva, Rodrigo Benenson, Jan Hosang, Matthias Hein, and Bernt Schiele.
\newblock Simple does it: Weakly supervised instance and semantic segmentation.
\newblock In {\em The IEEE Conference on Computer Vision and Pattern
  Recognition (CVPR)}, July 2017.

\bibitem{lensink2019fully}
Keegan Lensink, Eldad Haber, and Bas Peters.
\newblock Fully hyperbolic convolutional neural networks.
\newblock {\em arXiv preprint arXiv:1905.10484}, 2019.

\bibitem{JMLR:v11:mann10a}
Gideon~S. Mann and Andrew McCallum.
\newblock Generalized expectation criteria for semi-supervised learning with
  weakly labeled data.
\newblock {\em Journal of Machine Learning Research}, 11(32):955--984, 2010.

\bibitem{marquez2017imposing}
Pablo M{\'a}rquez-Neila, Mathieu Salzmann, and Pascal Fua.
\newblock Imposing hard constraints on deep networks: Promises and limitations.
\newblock {\em arXiv preprint arXiv:1706.02025}, 2017.

\bibitem{NIPS2019_9385}
Yatin Nandwani, Abhishek Pathak, Mausam, and Parag Singla.
\newblock A primal dual formulation for deep learning with constraints.
\newblock In H. Wallach, H. Larochelle, A. Beygelzimer, F. d'Alch\'{e} Buc, E.
  Fox, and R. Garnett, editors, {\em Advances in Neural Information Processing
  Systems 32}, pages 12157--12168. Curran Associates, Inc., 2019.

\bibitem{10.1109/ICCV.2015.203}
George Papandreou, Liang-Chieh Chen, Kevin~P. Murphy, and Alan~L. Yuille.
\newblock Weakly-and semi-supervised learning of a deep convolutional network
  for semantic image segmentation.
\newblock In {\em Proceedings of the 2015 IEEE International Conference on
  Computer Vision (ICCV)}, ICCV ?15, page 1742?1750, USA, 2015. IEEE Computer
  Society.

\bibitem{Pathak_2015_ICCV}
Deepak Pathak, Philipp Krahenbuhl, and Trevor Darrell.
\newblock Constrained convolutional neural networks for weakly supervised
  segmentation.
\newblock In {\em The IEEE International Conference on Computer Vision (ICCV)},
  December 2015.

\bibitem{IEEEDavisDataset}
F. {Perazzi}, J. {Pont-Tuset}, B. {McWilliams}, L.~V. {Gool}, M. {Gross}, and
  A. {Sorkine-Hornung}.
\newblock A benchmark dataset and evaluation methodology for video object
  segmentation.
\newblock In {\em 2016 IEEE Conference on Computer Vision and Pattern
  Recognition (CVPR)}, pages 724--732, June 2016.

\bibitem{doi:10.1190/INT-2018-0225.1}
Bas Peters, Justin Granek, and Eldad Haber.
\newblock Multiresolution neural networks for tracking seismic horizons from
  few training images.
\newblock {\em Interpretation}, 7(3):SE201--SE213, 2019.

\bibitem{peters2019symmetric}
Bas Peters, Eldad Haber, and Keegan Lensink.
\newblock Symmetric block-low-rank layers for fully reversible multilevel
  neural networks.
\newblock {\em arXiv preprint arXiv:1912.12137}, 2019.

\bibitem{FRHyperspectral}
Bas Peters, Eldad Haber, and Keegan Lensink.
\newblock Fully reversible neural networks for large-scale surface and
  sub-surface characterization via remote sensing.
\newblock In {\em AI for Earth Sciences workshop, International Conference on
  Learning Representations (ICLR)}. 2020.

\bibitem{peters2019algorithms}
Bas Peters and Felix~J Herrmann.
\newblock Algorithms and software for projections onto intersections of convex
  and non-convex sets with applications to inverse problems.
\newblock {\em arXiv preprint arXiv:1902.09699}, 2019.

\bibitem{10.5555/2969644.2969708}
John~C. Platt and Alan~H. Barr.
\newblock Constrained differential optimization.
\newblock In {\em Proceedings of the 1987 International Conference on Neural
  Information Processing Systems}, NIPS’87, page 612–621, Cambridge, MA,
  USA, 1987. MIT Press.

\bibitem{rajchl2016deepcut}
Martin Rajchl, Matthew~CH Lee, Ozan Oktay, Konstantinos Kamnitsas, Jonathan
  Passerat-Palmbach, Wenjia Bai, Mellisa Damodaram, Mary~A Rutherford, Joseph~V
  Hajnal, Bernhard Kainz, et~al.
\newblock Deepcut: Object segmentation from bounding box annotations using
  convolutional neural networks.
\newblock {\em IEEE transactions on medical imaging}, 36(2):674--683, 2016.

\bibitem{wateratlas}
Stanley~G. Robson and Edward~R. Banta.
\newblock Ground water atlas of the united states: Segment 2, arizona,
  colorado, new mexico, utah.
\newblock Technical report, U.S. Geological Survey, 1995.

\bibitem{10.1007/978-3-319-46484-8_25}
Fatemehsadat Saleh, Mohammad~Sadegh Aliakbarian, Mathieu Salzmann, Lars
  Petersson, Stephen Gould, and Jose~M. Alvarez.
\newblock Built-in foreground/background prior for weakly-supervised semantic
  segmentation.
\newblock In Bastian Leibe, Jiri Matas, Nicu Sebe, and Max Welling, editors,
  {\em Computer Vision -- ECCV 2016}, pages 413--432, Cham, 2016. Springer
  International Publishing.

\end{thebibliography}
}
\end{document}